\documentclass[conference]{IEEEtran}
\usepackage{cite}
\usepackage{amsmath,amssymb,amsfonts}
\usepackage{graphicx}
\usepackage{textcomp}
\usepackage{algpseudocode}
\usepackage{algorithm}
\usepackage{caption} 
\usepackage{hyperref}

\usepackage{float}
\usepackage{booktabs, multirow} 
\usepackage{multicol}
\usepackage{soul}
\usepackage[table]{xcolor} 

\begin{document}

\title{DEFER: Distributed Edge Inference for Deep Neural Networks}

\author{
\IEEEauthorblockN{ Arjun Parthasarathy
}
\IEEEauthorblockA{ Crystal Springs Uplands School \\
    \textit{Email: aparthasarathy23@csus.org}}
\and
\IEEEauthorblockN{Bhaskar Krishnamachari}
\IEEEauthorblockA{University of Southern California \\
    \textit{Email: bkrishna@usc.edu}}

}

\date{November 2021}

\maketitle
\begin{abstract}
  Modern machine learning tools such as deep neural networks (DNNs) are playing a revolutionary role in many fields such as natural language processing, computer vision, and the internet of things. Once they are trained, deep learning models can be deployed on edge computers to perform classification and prediction on real-time data for these applications. Particularly for large models, the limited computational and memory resources on a single edge device can become the throughput bottleneck for an inference pipeline. To increase throughput and decrease per-device compute load, we present DEFER (Distributed Edge inFERence), a framework for distributed edge inference, which partitions deep neural networks into layers that can be spread across multiple compute nodes. The architecture consists of a single “dispatcher” node to distribute DNN partitions and inference data to respective compute nodes. The compute nodes are connected in a series pattern where each node’s computed result is relayed to the subsequent node. The result is then returned to the Dispatcher. 
  We quantify the throughput, energy consumption, network payload, and overhead for our framework under realistic network conditions using the CORE network emulator. 
  We find that for the ResNet50 model, the inference throughput of DEFER with 8 compute nodes is 53\% higher and per node energy consumption is 63\% lower than single device inference. We further reduce network communication demands and energy consumption using the ZFP serialization and LZ4 compression algorithms.
  We have implemented DEFER in Python using the TensorFlow and Keras ML libraries, and have released DEFER as an open-source framework to benefit the research community.
\end{abstract}

\section{Introduction}

Machine Learning today is happening more than ever in real-world environments, and with the rise of the Internet of Things, or IoT, it is done with real-time sensor data. Historically, inference in this case would be performed in the cloud, by sending data collected from all IoT devices at the edge to the data center. However, this is a high latency solution that also incurs significant bandwidth costs, and is thus not suited for many IoT applications. Therefore, researchers and practitioners have advocated performing inference at the edge, close to the IoT data sources~\cite{8763885}.

Particularly for large models, the limited computational and memory resources on a single edge device can become the throughput bottleneck for an inference pipeline. To address this problem, we propose a framework to sequentially \textit{partition} the model in question and distribute these smaller partitions across multiple edge devices, such that each device sends its computed inference result to a subsequent device.

Our framework, DEFER (Distributed Edge inFERence), handles this process of partitioning the model, distributing across resource-constrained devices, and performing distributed inference on the edge. We have implemented DEFER in software and evaluate it empirically using a network emulator, to understand its performance in terms of throughput, energy efficiency, and network overhead and payload.

The following are the key contributions of this work: 
\begin{itemize}
    \item We present a software architecture and implementation for running ML inference on distributed nodes, where each node runs a portion of the given model.
    \item We evaluate this implementation using the CORE network emulator with different numbers of partitions/nodes using the traditional single node model as a baseline.
    \item Specifically, we evaluate the following metrics: throughput, energy consumption, network overhead, and network payload.
    \item We show through our experiments that DEFER with 8 compute nodes on ResNet50 has 53\% higher throughput and 63\% lower per-node energy consumption compared to single device inference, and that communication demands are reduced with the LZ4 and ZFP algorithms.
    \item Our software implementation is open-sourced for the research community, and available at \url{https://github.com/ANRGUSC/DEFER}.
\end{itemize}

Below, we present pertinent literature, DEFER's architecture and implementation, methods of experimentation, and the results and conclusions of our research.

\section{Related Work}
Prior work on distributed machine learning can be broadly classified into two categories: \textit{distributed training} and \textit{distributed inference}. We briefly review the literature, noting that our work is focused on distributed inference.

\subsection{Distributed Training}
\subsubsection{Model Parallelism}
Prior work includes many frameworks to optimize communication demands and model partitioning~\cite{dean2012large, lin2020deep}. These frameworks introduce model parallelism for simultaneous execution and optimization of model layers. 

\subsubsection{Federated Learning} 
Federated Learning~\cite{9084352} deals with model training on restricted data over the cloud, in privacy-preserving applications with sensitive data. Unlike Distributed Training, it facilitates local data sharing rather than reducing training time.

\subsection{Distributed Inference}

\subsubsection{Hybrid Computation}
Frameworks such as MCDNN~\cite{Shen2015MCDNNAE} and DDNN~\cite{teerapittayanon2017distributed} propose DNN size optimization to run jointly on-device and on the cloud, with a small tradeoff in model accuracy. Unlike these works, we do not alter the architecture of the original model to be run in a different setting; rather we partition an existing model across multiple edge devices without any cloud processing.

\subsubsection{Layer-Wise Fusing and  Partitioning}

Prior works~\cite{8493499, matsubara2019distilled, Abdi2020RestructuringPA, Hadidi2020LCPAL, 9097616, 10.1145/3404397.3404473, 10.1145/3386367.3431666} have considered partitioning the model across 2 or more devices; they have
proposed different architectures to optimize both Dense and Convolutional layers, such as \textit{FTP} (Fused-Tile Partitioning), \textit{FDSP} (Fully Decomposable Spatial Partition) and \textit{Network Distillation}.
Unlike our architecture which preserves the exact configuration and inference accuracy of the original model, these works approximate or modify layers to reduce communication and computation demand, at the cost of slightly reduced model accuracy. In principle, the techniques described in these works are complementary as our framework could be extended to use the modified/distilled models proposed by these works.

\subsubsection{Software Implementations}

We also briefly review related works that are not published research papers but have published software implementations.
\begin{itemize}
    \item  ``Train and Pruning with PyTorch"~\cite{GHPrune} - Implements layer pruning for model size reduction, as seen in the \textit{Layer-Wise Pruning and Partitioning} section. This method introduces a loss in model accuracy, which we avoid with our framework.
    \item ``DistributedInference"~\cite{KennyDist} - Distributes a copy of the model to each worker node, and sends inference jobs to a free node, thereby computing multiple inference jobs at once. This approach introduces complete inference parallelization. Unlike ours, this project does not address system resource limitations of individual edge devices.
    \item ``Distributed Inference Platform"~\cite{MelDist} - Also performs parallelized inference on exact copies of the model, and takes advantage of existing edge inference frameworks, such as Google EDGETPU and Nvidia TensorRT. This project does not address system resource limitations of individual edge devices.
\end{itemize}

\section{DEFER Architecture and Implementation}

\begin{figure}[htbp]
\centerline{\includegraphics[scale=0.1]{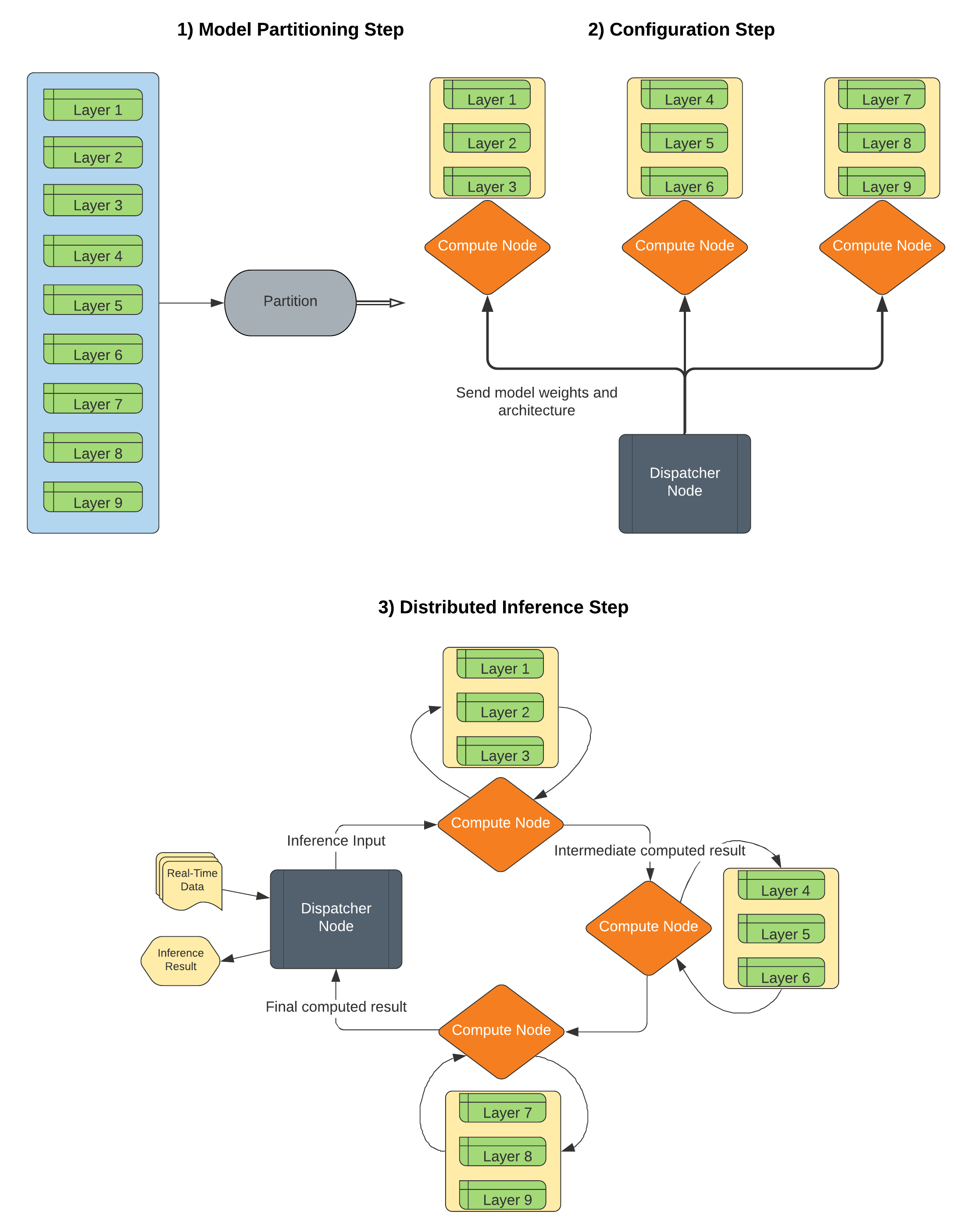}}
\caption{Architecture of DEFER}
\label{fig:architecture}
\end{figure}

DEFER is made up of a central \textit{Dispatcher Node}, which creates and distributes partitioned models, and multiple \textit{Compute Nodes}, which run inference calculations with their respective partition.

DEFER's implementation can be split up into three steps, the \textit{Model Partitioning Step}, the \textit{Configuration Step}, and the \textit{Distributed Inference Step}.

\subsection{Model Partitioning Step}

Partitioning the model involves splitting it layer-wise into multiple sequential sub-networks, as illustrated in Figure \ref{fig:architecture}.
To implement model partitioning in Python, we used the TensorFlow~\cite{tensorflow2015-whitepaper} and Keras~\cite{chollet2015keras} libraries. We took advantage of the DAG (\textit{Directed Acyclic Graph}) structure of Keras Layers to create a graph traversal algorithm, which traverses the section of the DAG that we want to partition and produces a new DAG with the desired layers. Partitioning can be done with any layer graph configuration, not just with sequential layers as shown in Figure \ref{fig:architecture}. A new model of just the partitioned layers is created, which can then be sent to the compute nodes.

\subsection{Configuration Step}
To setup the partitions on the compute nodes, we first define our dispatcher node run-time and our compute node run-time. The dispatcher node connects to two TCP Sockets for each compute node. One TCP Socket is used to send a serialized representation of the model's architecture and the IP address of the next node in the inference chain, while the other sends a serialized array of the model's weights. When receiving the model on the compute node end of the socket, the model socket waits for the model's weights to be received on the other socket, upon which time it instantiates a full TensorFlow model with the correct architecture and weights.

On a separate thread, each compute node opens a TCP socket connection to its assigned next node, which will be used to send intermediate inference results to the subsequent node in the inference chain. This results in each node having an incoming socket data connection from the previous node and an outgoing connection to the following node.

\subsection{Distributed Inference Step}
The dispatcher, having distributed model partitions across compute nodes, now begins to send data to the first compute node in the chain. Upon receiving data on its incoming socket, the node performs the necessary de-serialization and compression and runs that previous inference result through its model. The result is then serialized and compressed and sent via the outgoing socket. This process is repeated for each compute node.

By nature, the sockets compute inference calculations in a serial first-in first-out manner, whereby prior inference data will be sent to the next node earlier. This is done to maintain the order in which inference results are returned to the Dispatcher Node. Since a compute node can take in new inference data after it has finished a prior inference calculation rather than waiting for the entire model to finish inference, the system will have higher inference throughput.

We employ chunked data transfer (with a default size of $512 kB$ per chunk) across all of our socket connections, due to the high volume of information required to construct a model and send intermediate inference results.
\bigbreak
We present the pseudocode for our implementation in algorithms \ref{alg:dispatcher} and \ref{alg:compute} describing the dispatcher node and compute nodes, respectively. 

\begin{algorithm}
\caption{Dispatcher Node}\label{alg:dispatcher}
\begin{algorithmic}
\algblockdefx{Spawn}{EndSpawn}[1]{\textbf{spawn} #1}{}
\State{\textbf{//} Using DAG traversal}
\State{$partitions \gets \Call{PARTITION-MODEL}$}
\\
\State{\textbf{//} Configuration Step}
\For {$i \gets 1 \textbf{ to } partitions.\Call{length}$}
    \State{\textbf{//} Serialization and compression}
    \State{$wSock \gets \Call{OPEN-SOCKET}{\text{weightsPort, nodes[i]}}$}
    \State{$mSock \gets \Call{OPEN-SOCKET}{\text{modelPort, nodes[i]}}$}
    \State {$wSock.\Call{SEND}{partitions[i].weights}$}
    \State {$mSock.\Call{SEND}{partitions[i].architecture}$}
    \State{\textbf{//} Tell node which node to connect to in the chain}
    \State {$mSock.\Call{SEND}{nodes[i+1]}$}
\EndFor
\\
\State{\textbf{//} Distributed Inference Step}
\State{$nodeSock \gets \Call{OPEN-SOCKET}{\text{firstNodeIP}}$}
\Spawn{THREAD-1}
    \Repeat
    \State{\textbf{//} Serialized and compressed}
    \State {$toInference \gets \textbf{pipe from } \text{MAIN-THREAD}$}
    \State {$nodeSock.\Call{SEND}{toInference}$}
    \Until{SYSEXIT}
\EndSpawn
\Spawn{THREAD-2}
    \State{$outServer.\Call{LISTEN}$}
    \Repeat
    \State{$output = outServer.\Call{READ}$}
    \State{$\textbf{pipe } output \to \text{MAIN-THREAD}$}
    \Until{SYSEXIT}
\EndSpawn
\end{algorithmic}
\end{algorithm}

\begin{algorithm}
\caption{Compute Node}\label{alg:compute}
\begin{algorithmic}
\algblockdefx{Spawn}{EndSpawn}[1]{\textbf{spawn} #1}{}
\Spawn{WEIGHTS-THREAD}
\State{$wServer.\Call{LISTEN}$}
\State{\textbf{//} Deserialization and decompression}
\State{$weights \gets sockServer.\Call{READ}$}
\State{$\textbf{pipe } weights \to \text{MODEL-THREAD}$}
\EndSpawn
\Spawn{MODEL-THREAD}
\State{$mServer.\Call{LISTEN}$}
\State{$mConfig \gets mServer.\Call{READ}$}
\State{$toDataNode  \gets mServer.\Call{READ}$}
\State{$weights \gets \textbf{pipe from } \text{MODEL-THREAD}$}
\State{$model \gets \Call{MAKE-MODEL}{weights, mConfig}$}
\State{$toDataSock \gets \Call{OPEN-SOCKET}{toDataNode}$}
\EndSpawn
\State{\textbf{//} Separate threads for reading and sending data
to avoid inference bottleneck}
\Spawn{THREAD-1}
    \Repeat
    \State{\textbf{//} Inference result from previous node}
    \State{$data \gets \Call{READ-DATA}$}
    \State{$\textbf{pipe } data \to \text{THREAD-2}$}
    \Until{SYSEXIT}
\EndSpawn
\Spawn{THREAD-2}
    \Repeat
    \State{\textbf{//} Compute inference with model and data}
    \State{$\textbf{pipe from } \text{THREAD-1} $}
    \State{$result \gets \Call{INFERENCE}{model, data}$}
    \State{$toDataSock.\Call{SEND}{result}$}
    \Until{SYSEXIT}
\EndSpawn
\end{algorithmic}
\end{algorithm}

\section{Evaluation Methodology}

We used the VGG-16 and VGG-19~\cite{simonyan2014very} and ResNet50~\cite{He_2016_CVPR} models, all trained on the ImageNet dataset, to evaluate our metrics. These models were chosen because they are widely used in computer vision and contain enough layers to be used with DEFER. The partitioning layers were selected based on what would split the model up into a similar number of layers for each partition. 
For each model, we varied the number of compute nodes between 4, 6, and 8. This was done using the CORE Network Emulator (\url{http://coreemu.github.io/core}), which allowed us to create network topologies simulating these different node configurations and run the simulation locally in a close-to-zero latency environment.

With a configuration of ResNet50 with 4 compute nodes, we used different compression and serialization schemes for each of the three sockets, which transfer the model architecture and weights during the configuration step and inference data during the inference step, respectively. With these combinations, we computed different metrics, namely inference throughput, network-related energy consumption (the energy expended both serializing/compressing data and sending the data over the network), overhead, and network payload.

\textbf{Serialization/Compression Algorithms:} To serialize our data, we tested both JSON serialization of NumPy arrays and ZFP~\cite{article}. We further tested both serialization configurations with and without the use of the LZ4 compression algorithm.

\textbf{Inference Throughput:} To measure inference throughput, we set a fixed time of execution for DEFER, and recorded how many inference cycles could be done in that fixed time. This gave us a measure of throughput in inference cycles per second.

\textbf{Energy Consumption:} To measure energy consumption for each different serialization algorithm, we calculated the amount of time it took the given algorithms to run and multiplied that by the TDP (Thermal Design Power). Additionally, we measured the energy consumption of sending the data over the network, which is proportional to the network payload and varies between communication medium. For our tests, we used metrics associated with an Ethernet connection ($10pJ$ per bit~\cite{bit_energy}).

\textbf{Overhead:} To measure overhead, we recorded the amount of time spent formatting data to be sent over the network.

\textbf{Network Payload:} To measure network payload, we used a network tool called nload~\cite{DriegelNload}, which gave us the total amount of data sent over the TCP sockets.

\section{Results}

\begin{figure}[htbp]
\centerline{\includegraphics[scale=0.4]{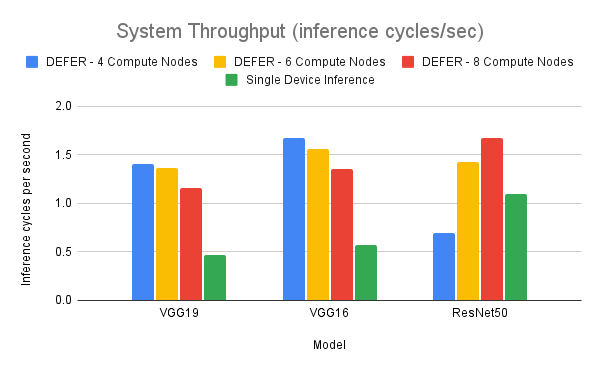}}
\caption{Inference Throughput for different models and number of Compute Nodes}
\label{fig:comparison}
\end{figure}

According to Figure \ref{fig:comparison}, DEFER outperforms single device inference in terms of throughput, for various numbers of compute nodes used. It is apparent, however, that there is a limit to an increase in throughput from utilizing additional compute nodes, beyond which the total time of overhead will outweigh the gains from parallelization. For example, increasing the number of partitions decreases the resulting throughput of the VGG16 model, but due to ResNet50's size, each partition for that model is sufficiently large whereby network overhead is comparatively trivial.

\begin{table*}[!htp]\centering
\caption{Energy Consumption, Overhead, and Network Payload for ResNet50 with 4 Compute Nodes}\label{tab:metrics}
\scriptsize
\begin{tabular}{lllrrr}\toprule
Type &Serialization &Compression &Energy Consumption (J) &Overhead (s) &Network Payload (MB) \\\midrule
Architecture &JSON &LZ4 &0.019 &4.17e-4 &0.02389 \\
\midrule
Architecture &JSON &Uncompressed &0.0032 &1.45e-5 &0.13035 \\
\midrule
Weights &JSON &LZ4 &4.4671 &19.4702282 &446.7 \\
\midrule
Weights &JSON &Uncompressed &5.5166 &8.326519966 &551.66 \\
\midrule
Weights &ZFP &LZ4 &3.0933 &16.33518434 &309.32 \\
\midrule
Weights &ZFP &Uncompressed &5.1283 &14.49310565 &512.83 \\
\midrule
Data &JSON &LZ4 &0.1294 &0.4661076069 &12.939 \\
\midrule
Data &JSON &Uncompressed &0.1754 &0.415356636 &17.543 \\
\midrule
Data &ZFP &LZ4 &0.1051 &0.3867843151 &10.513 \\
\midrule
Data &ZFP &Uncompressed &0.1423 &0.3257746696 &14.233 \\
\bottomrule
\end{tabular}
\end{table*}

Table \ref{tab:metrics} compares different serialization and compression configurations and the metrics associated with each.

We see for the dispatcher to communicate the model architecture to the compute nodes, JSON without compression is the best communication configuration, because the JSON representation of the architecture is small enough such that the computational trade-off from using compression outweighs the reduction in network payload. 

For the dispatcher to communicate the weights to the compute nodes, using ZFP and LZ4 was the best configuration. For large NumPy arrays such as the model's weights array, the difference in the size of the serialized representations from ZFP and JSON becomes more apparent. Furthermore, the almost 25\% reduction in network payload from using LZ4 suggests it outweighs the additional computational cost.

For communicating data between nodes in the distributed inference step, ZFP and LZ4 was also the best configuration. The use of NumPy arrays to transfer the data warrants ZFP instead of JSON serialization, just like the weights array.

\begin{table}[htbp]\centering
\caption{Inference Throughput for Different Serialization and Compression Configurations}\label{tab:throughput}
\scriptsize
\begin{tabular}{llr}\toprule
Serialization &Compression &Inference Throughput (cycles/sec) \\\midrule
JSON &LZ4 &0.477 \\
\midrule
JSON &Uncompressed &0.493 \\
\midrule
ZFP &LZ4 &0.673 \\
\midrule
ZFP &Uncompressed &0.5 \\
\bottomrule
\end{tabular}
\end{table}

Table \ref{tab:throughput} shows that ZFP with LZ4 results in the highest inference throughput. When running DEFER as a high-volume system, communication demands become increasingly important, and using ZFP with LZ4 minimizes the amount of data sent over the network and facilitates throughput, despite the additional computational cost.

\begin{figure}[h!]
\centerline{\includegraphics[scale=0.3]{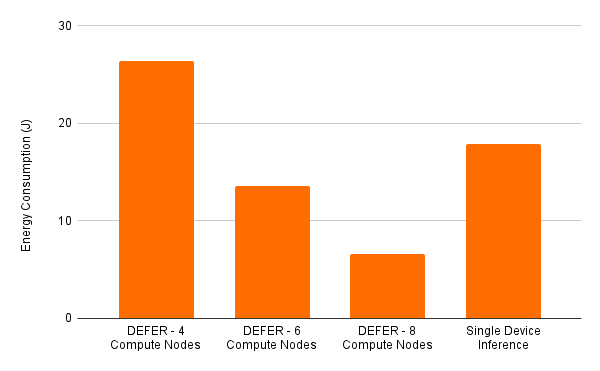}}
\caption{Energy consumption per node for DEFER and Single Device Inference, using ResNet50}
\label{fig:pernode}
\end{figure}

Figure \ref{fig:pernode} describes the average energy consumption per node for an inference cycle. As the number of nodes increases, the per node consumption decreases and falls below the amount for single device inference. This pattern coincides with the corresponding increase in throughput for higher numbers of compute nodes found in Figure \ref{fig:comparison}. With 4 compute nodes, each partition is sufficiently large such that computational requirements create large per-node energy consumption. Figure \ref{fig:pernode} shows that at least 6 compute nodes are necessary for a per-node energy consumption that is lower than single device inference.

\section{Conclusions}

We have presented DEFER, a framework for Distributed Inference on resource-constrained edge devices, via an algorithm that performs DNN partitioning. These partitions are sent to their respective compute nodes, and used to perform inference. We have shown that DEFER with 8 Compute Nodes outperforms Single Device Inference in terms of inference throughput by 53\%, and has a 63\% lower average energy consumption across nodes. We further found that overhead and energy consumption can be reduced with the use of the ZFP serialization and LZ4 compression algorithms. DEFER is available to the research community as an open-source framework at \url{https://github.com/ANRGUSC/DEFER}.

In future work on DEFER, we plan to explore how to optimize model partition size and architecture based on the compute and memory constraints of the edge device. In a situation without homogeneous compute nodes, heterogeneous model partitions can be more effectively distributed for higher inference throughput. Furthermore, we can introduce the concept of a \textit{virtual node}, where multiple partitions are dynamically loaded into and run on a single physical compute node. Through this process, partitions may have multiple inputs and outputs, and DEFER's code will optimized to support this architecture.

\bibliography{citations}
\bibliographystyle{ieeetr}
\end{document}